\def\year{2020}\relax
\documentclass[letterpaper]{article}
\usepackage{aaai2020}
\usepackage{times}
\usepackage{helvet}
\usepackage{courier}
\usepackage{graphicx}
\usepackage{booktabs}
\usepackage{multirow}
\usepackage{amssymb}
\usepackage{subfigure}
\usepackage{color}
\usepackage{amsmath}

\usepackage[hyphens]{url}  
\begin{document}
%
\title{Grapy-ML: Graph Pyramid Mutual Learning for Cross-dataset Human Parsing}
\author{
Haoyu He,\textsuperscript{ \thanks{equal contribution.}}
Jing Zhang,\textsuperscript{ \footnotemark[1]}
Qiming Zhang, \ 
Dacheng Tao
\\
UBTECH Sydney AI Centre, School of Computer Science, \\ Faculty of Engineering, The University of Sydney, Darlington, NSW 2008, Australia \\
\{hahe7688, qzha250\}@uni.sydney.edu.au, \{jing.zhang1,dacheng.tao \}@sydney.edu.au
}
\maketitle

\begin{abstract}
Human parsing, or human body part semantic segmentation, has been an active research topic due to its wide potential applications. In this paper, we propose a novel GRAph PYramid Mutual Learning (Grapy-ML) method to address the cross-dataset human parsing problem, where the annotations are at different granularities. Starting from the prior knowledge of the human body hierarchical structure, we devise a graph pyramid module (GPM) by stacking three levels of graph structures from coarse granularity to fine granularity subsequently. At each level, GPM utilizes the self-attention mechanism to model the correlations between context nodes. Then, it adopts a top-down mechanism to progressively refine the hierarchical features through all the levels. GPM also enables efficient mutual learning. Specifically, the network weights of the first two levels are shared to exchange the learned coarse-granularity information across different datasets. By making use of the multi-granularity labels, Grapy-ML learns a more discriminative feature representation and achieves state-of-the-art performance, which is demonstrated by extensive experiments on the three popular benchmarks, $e.g.$ CIHP dataset. The source code is publicly available at https://github.com/Charleshhy/Grapy-ML.
\end{abstract}

\section{Introduction}
Human parsing, or human body part semantic segmentation, refers to assigning dense pixel-wise category labels for each human body parts. It is a fundamental computer vision task and plays a critical role in human-centric analysis and potential down-stream applications, $e.g.$, action recognition, video surveillance and virtual reality. However, it is very challenging due to the large appearance variance, the lack of quality on labeled samples and the domain gap between training and testing data.

\begin{figure}
\centering
\includegraphics[width=.95\linewidth]{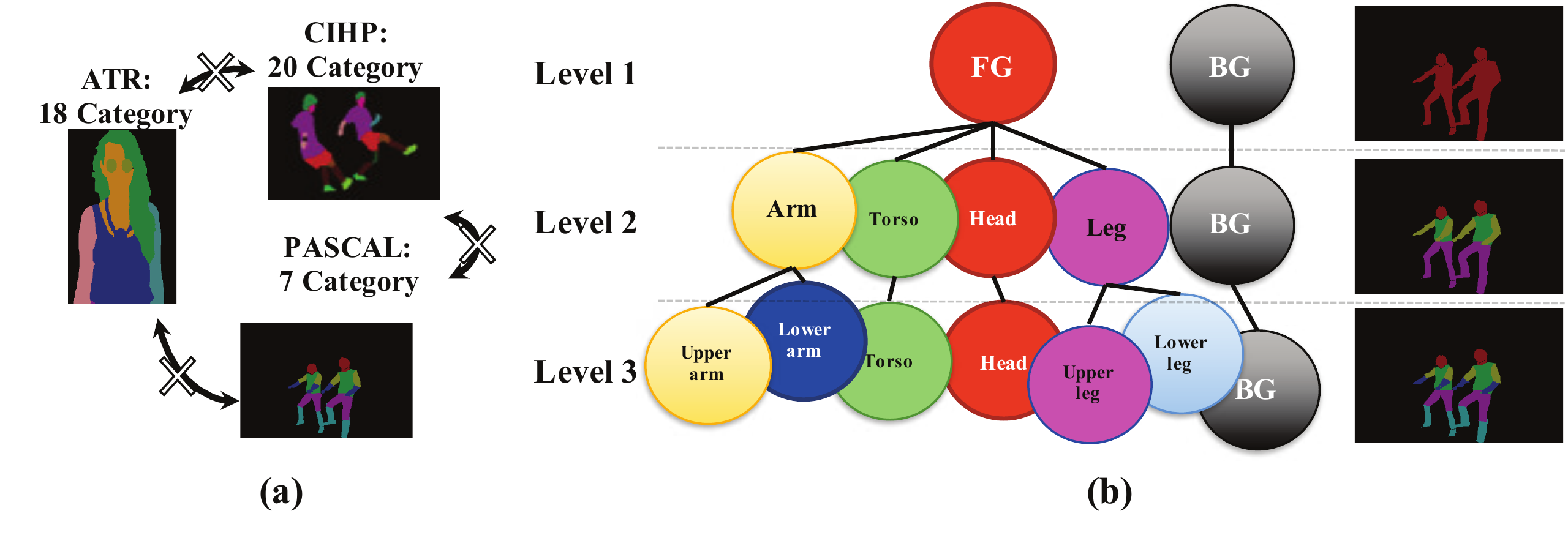}
\caption{(a) Label discrepancy between different datasets. (b) Multi-granularity lexical pyramid representation of the human body.}
\label{fig:labelGranularity}
\end{figure}

To deal with these issues, prior methods have achieved significant progress based on the success of the deep neural network. Since human body is highly structural, many methods have been proposed to model the context correlations efficiently between body parts using convolutional neural network (CNN) \cite{liang2015human,li2017multiple,zhu2018progressive,ruan2019devil}, recurrent network with long short-term memory units (LSTM) \cite{liang2016semantic,liang2016semantic,liang2017interpretable}, and graph convolutional neural network (GCN) \cite{gong2019graphonomy}. For instance, Ruan $et~al.$ propose a context embedding with edge perceiving (CE2P) network for single human parsing after identifying three key factors affecting the parsing performance, including high-resolution maintenance, global context embedding, and edge perceiving \cite{ruan2019devil}. However, multi-human parsing is a more common setting for real-world applications. To this end, Li $et~al.$ introduce a multi-human parsing (MHP) dataset and a novel bottom-up multi-human parser based on graph-GAN model \cite{li2017multiple}. Besides, many researches use LSTM to model structural dependencies in feature learning, such as local-global LSTM \cite{liang2016semantic}, Graph LSTM \cite{liang2016semantic2}, and structure-evolving LSTM \cite{liang2017interpretable}. However, these methods do not utilize the strong hierarchical relations in body parts. In contrast, Zhu $et~al.$ propose a progressive cognitive network (PCNet) to segment hierarchical human parts using a component-aware region convolution structure \cite{zhu2018progressive}. Although progressive cognitive human parsing achieves better results than segmenting each part simultaneously, it still remains open that how to efficiently incorporate the hierarchically structural prior information into feature learning to further improve the parsing performance.

Another direction is to explore multiple datasets or heterogeneous annotations by utilizing multi-task learning, transfer learning, or mutual learning \cite{xiao2018unified,nie2018mutual,gong2019graphonomy}. For instance, Xiao $et~al.$ propose a multi-task learning framework to learn visual concepts from heterogeneous image annotations for unified perceptual scene parsing \cite{xiao2018unified}. However, adding extra parallel branches for different tasks does not explicitly model their relationships. Nie $et~al.$ propose a mutual learning method adapting human parsing and pose estimation, which explicitly incorporates the guidance information from their parallel tasks via a mutual adaptation module \cite{nie2018mutual}. As shown in Figure~\ref{fig:labelGranularity}(a), the annotations from different human parsing datasets are at different granularity levels. It is not trivial to convert them into a consensus form or transfer one to another. To leverage the granularity-inconsistency problems, Gong $et~al.$ propose a universal human parsing model named Graphonomy by integrating intra-graph reasoning and inter-graph transferring together \cite{gong2019graphonomy}. A learnable transfer matrix is utilized to directly map the heterogeneous graph representation from one dataset to another. Although the transfer dependency between source graph nodes and target graph nodes is enabled, the transfer matrix is strictly defined by the granularity level of annotations, for example, a $R^{7\times20}$ matrix from PASCAL-Person Part dataset to CIHP dataset. 

In fact, as shown in Figure~\ref{fig:labelGranularity}(b), the heterogeneous annotations at different granularity levels still share some underlying coarse-granularity concepts, which can be explored to improve the transferring performance further. Motivated by this observation, we propose a novel graph pyramid mutual learning (Grapy-ML) method to address the cross-dataset human parsing problem by making use of the heterogeneous multi-granularity annotations. At first, we define two levels of coarse-granularity categories as shown in the upper part of Figure~\ref{fig:labelGranularity}(b). At the coarsest level, only the foreground human body and the background are involved. Then, we divide the foreground human body into four parts with clear semantics, $e.g.$, head, torso, legs and arms. As can be seen, Although the annotations from different benchmarks are diverse, they share the same underlying coarse-level categories as defined above. Based on the definition, we devise a graph pyramid module (GPM) by stacking three levels of graph structures from coarse granularity to fine granularity subsequently. At each level, the node represents a category-wise feature representation by aggregating the pixel-level encoding features from the previous stage according to the category mask. GPM utilizes the self-attention mechanism to model the correlations between context nodes. Then, it adopts a top-down mechanism to progressively refine the hierarchical node features through all the levels. In this way, the proposed GPM is able to explicitly incorporate the hierarchical structural prior to feature learning. 

Moreover, GPM also enables efficient mutual learning owing to its hierarchical multi-granularity structure. Since the annotations from different datasets share the same coarse-level categories, we keep the network weights of the first two levels shared across different datasets to learn a common coarse-granularity category-wise feature representation. Consequently, due to the accessibility of more training samples, GPM is able to learn more robust coarse-level features, which further enhances the finest-level features progressively with the hierarchical refinement and improve the final parsing performance correspondingly.

The main contributions of this paper are summarized as follows. Firstly, we define a hierarchical multi-granularity representation of the human body. Based on it, we propose a novel graph pyramid module, which enables incorporating the hierarchical structural prior explicitly into feature learning via self-attention based graph reasoning and progressive feature refinement. Secondly, we build a novel mutual learning method on the GPM, which leverages the multi-granularity annotations explicitly from different human parsing datasets. Finally, the proposed Grapy-ML model achieves state-of-the-art performance on several popular benchmarks including PASCAL-Person Part dataset, CIHP dataset, and ATR dataset.

\section{Related work}
Human parsing methods can be categorized into single human parsing methods \cite{liang2015human,gong2017look} and multi-human parsing methods \cite{li2017multiple,gong2018instance,zhao2018understanding}. It is noteworthy that single human parsing methods can also be used for multi-human parsing after integrating with a person detector \cite{he2017mask}, known as instance-level multi-human parsing \cite{ruan2019devil,yang2019parsing}. Our method belongs to the latter category which segments all body parts within a given image. Since it mainly targets at improving the learned model from multiple datasets with heterogeneous multi-granularity annotations, we briefly review recent methods related to ours from the following two aspects: 1) hierarchical structure modeling; 2) multi-task learning and transfer learning.

The human body is highly structural which has definite semantical body parts, $e.g.$, arms, legs. Different body parts are connected by physical joints, $e.g.$, elbows and knees. This structural prior knowledge can be utilized during modeling and training for learning a better feature representation \cite{dong2014towards,dai2016instance,huang2017coarse,xiao2018unified}. For example, Nie $et~al.$ propose a mutual adaptation module to guide the human parsing task by leveraging the learned structural information from the pose estimation task and vice versa \cite{nie2018mutual}. Gong $et~al.$ propose a self-supervised structure-sensitive learning approach for human parsing, which generates pose structures from parsing results and impose joint structure loss as extra self-supervision \cite{gong2017look}. In addition, the human body has a hierarchical structure with different granularities. For example, it consists of upper and lower parts where the upper body has arms, neck, and head, and the arms can be further categorized as upper-arm and lower-arm. The hierarchical structure prior is also useful for improving parsing results. Recently, Zhu $et~al.$ propose to parse the human body progressively where layers benefit from prior coarse-granularity component information from the previous layers \cite{zhu2018progressive}. Luo $et~al.$ propose a macro-micro adversarial net to enforce low-level local consistency and high-level semantic consistency \cite{luo2018macro}. In contrast to them, we devise a graph pyramid module to explicitly model the hierarchical multi-granularity structure of the human body. It aims to progressively refine the learned features rather than progressively predicting the parsing results.

Recognizing a human body can be carried out at different granularities. For example, for PASCAL-Person part dataset, it only needs to parse 7 categories of body parts. However, for CIHP dataset, the parsing task should be carried out at all 20 categories. Due to the heterogeneous annotations from different datasets, it is straight forward to employ multi-task learning. For example, Gong $et~al.$ propose a part grouping network to unify the semantic part segmentation task and instance-aware edge detection. Xiao $et~al.$ propose a multi-task framework to learn from heterogeneous image annotations for unified perceptual parsing \cite{xiao2018unified}. The most related work to this paper is Graphonomy, which is proposed by Gong $et~al$ \cite{gong2019graphonomy}. It aims at conducting universal human parsing which explicitly utilizes multiple dataset heterogeneous annotations by intra-graph reasoning and inter-graph transfer and avoids fitting the parsing results into one dataset. There are several differences between Graphonomy and the proposed Grapy-ML: 1) we propose a graph pyramid module to explore the multi-granularity hierarchical structure of human body at each dataset. In contrast, they use a transfer matrix to explore the dependency between the graph representations predefined according to the annotations of each dataset; 2) we propose a mutual learning approach by sharing the first two coarse levels across different datasets, where they use inter-graph transfer to exchange the learned knowledge; 3) we use category-aware pooling to aggregate the semantics as the node features and self-attention for context information propagation. However, they directly project the image features into node features and use graph convolution to perform graph propagation. Experiments on three benchmarks validate the competitiveness of Grapy-ML against Graphonomy thanks to the GPM and mutual learning.

\section{Graph Pyramid Mutual Learning for Cross-dataset Human Parsing}
\label{sec:grapyml}

In this section, we introduce a novel graph pyramid mutual learning method to address the cross-dataset human parsing problem by leveraging multi-granularity annotations. Firstly, we define a hierarchical multi-granularity lexical representation of the human body based on its structural prior. Then, inheriting from this representation, we build up a graph pyramid module, which can be integrated into the deep encoder-decoder framework seamlessly. The GPM not only enables graph reasoning between context nodes but also enables mutual learning across different datasets. Details are presented as follows. 

\subsection{Hierarchical Multi-granularity Representation}
\label{subsec:hierarchicalDefinition}
As is shown in Figure~\ref{fig:labelGranularity}(b), the human body is highly structural and can be viewed at different granularities. For simplicity, we decompose the human body into three levels after analyzing the linguistic connections between different body parts at different granularities. For instance, \textbf{Level 1: Foreground and Background}, where the foreground refers the human body as a whole; \textbf{Level 2: Head, Torso, Arm, Leg, and Background}, where the human body is roughly divided into four sub-parts; \textbf{Level 3: Head, Torso, Upper Arm, Lower Arm, Upper Leg, Lower Leg, and Background}. Level 3 is specified by the exact definition of annotations in each dataset. Here we take PASCAL-Person Part dataset as an example. As can be seen, the two categories of Arm and Leg are divided into two finer sub-parts further. For CIHP dataset and ATR dataset, there are 18 and 20 categories at Level 3 correspondingly, $i.e.$, \textbf{Level 3 (CIHP): Face, Hat, Hair, Sunglasses, Upper Clothes, Dress, Coat, Socks, Pants, Torso Skin, Scarf, Skirt, Left Arm, Right Arm, Left Leg, Right Leg, Left Shoe, Right Shoe, and Background}, and \textbf{Level 3 (ATR): Face, Hat, Hair, Sunglasses, Upper Clothes, Dress, Pants, Scarf, Skirt, Belt, Bag, Left Arm, Right Arm, Left Leg, Right Leg, Left Shoe, Right Shoe, and Background}. It can be seen that all the categories at Level 3 within each dataset share the same coarse-granularity categories at Level 1 and Level 2.

\begin{figure}
\centering
\includegraphics[width=1\linewidth]{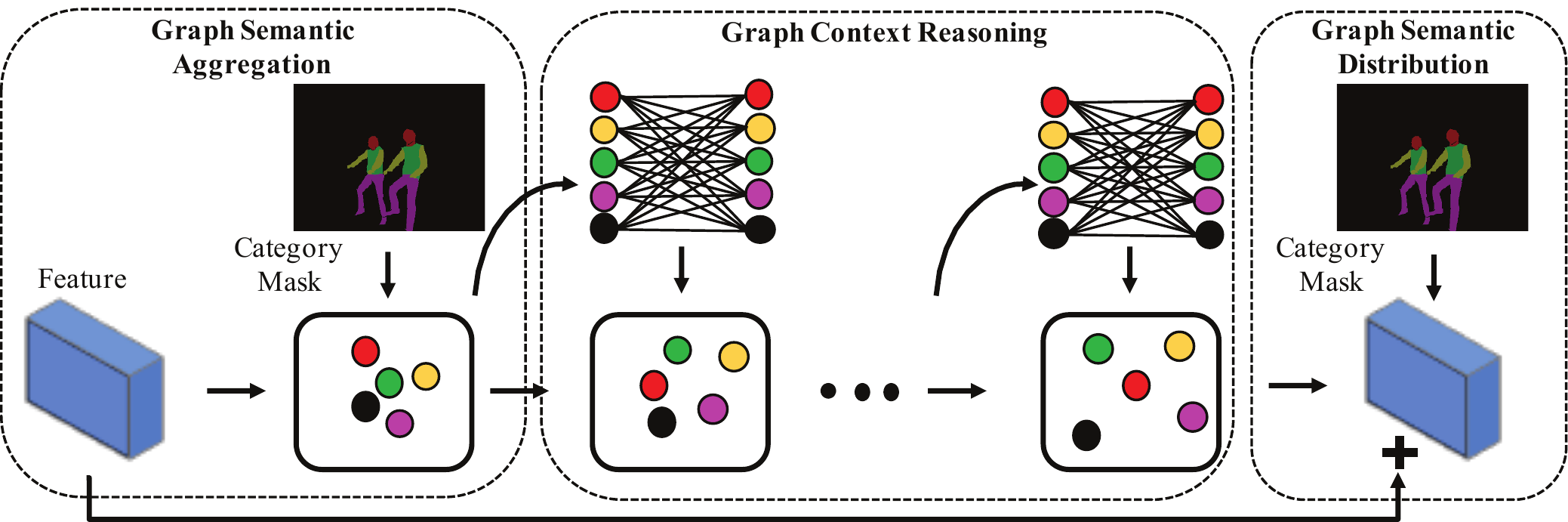}
\caption{Illustration of the proposed GPM at Level 2.}
\label{fig:gpmLevel2}
\end{figure}

\subsection{Graph Pyramid Module}
\label{subsec:GPM}
Based on the above definition, we devise a graph pyramid module by stacking three levels of graph structures from coarse granularity to fine granularity subsequently. At each level, GPM utilizes self-attention mechanism to model the correlations between context nodes and refine the learned features. Generally, it can be divided into three phases: Graph Semantics Aggregation (GSA), Graph Context Reasoning (GCR), and Graph Semantics Distribution (GSD) as illustrated in Figure~\ref{fig:gpmLevel2}. Details are presented as follows.

\begin{figure*}[ht]
\centering
\includegraphics[width=.95\linewidth]{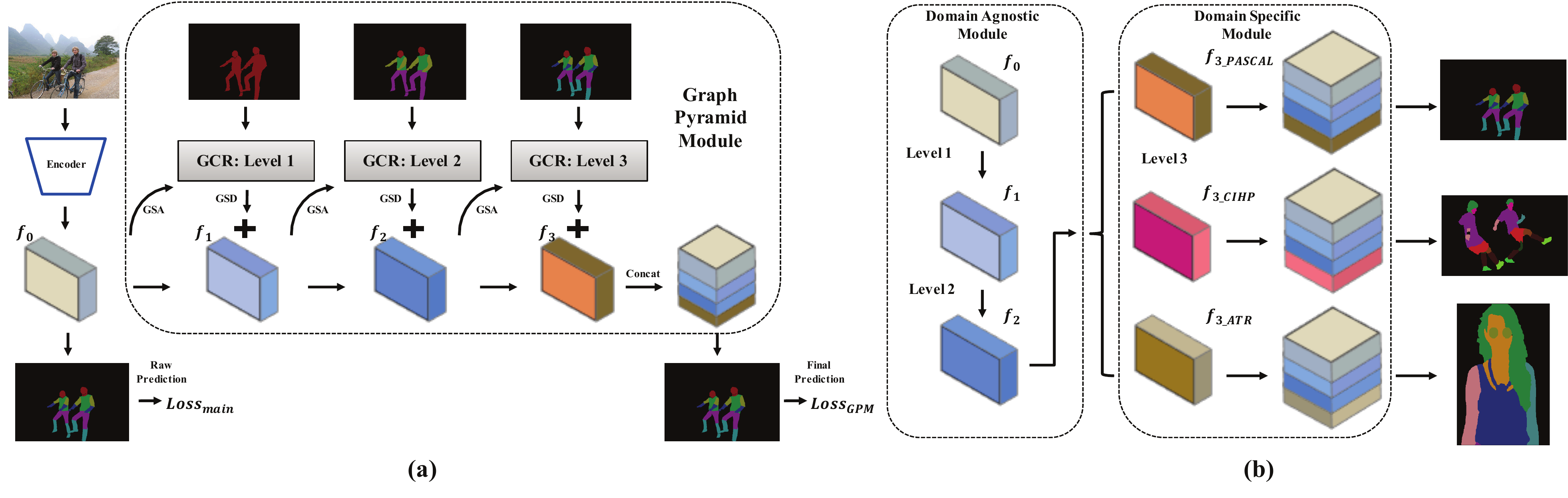}
\caption{Diagram of the proposed Grapy-ML. (a) The structure of the graph pyramid module. (b) The structure of the proposed GPM-based mutual learning. Please refer to the main text for more details.}
\label{fig:grapyml}
\end{figure*}

\subsubsection{Graph Semantics Aggregation}
\label{subsec:GSA}
In this paper, we choose Deeplab v3+ as the strong baseline and choose the decoded feature from the penultimate layer as the input of our GPM since these features are directly related to the final predicted categories. For simplicity, we denote the decoded feature as $f \in {R^{H \times W \times C}}$, where $h$, $w$, and $c$ denotes the height, width, and channels of the feature map. Then, the prediction by Deeplab v3+ can be written as:
\begin{equation}
  y = p\left( f \right),
\label{eq:rawprediction}
\end{equation}
where $p\left(  \cdot  \right)$ denotes the final prediction layer in Deeplab v3+, $y \in {R^{H \times W \times K}}$, $K$ denotes the number of parsing category.

To map the decoded feature into graph representation, we adopt category-aware pooling to aggregate the semantical features at each category, namely \emph{graph semantics aggregation}. Specifically, we utilize $y$ as the raw parsing results and calculate the category-wise feature as follows:
\begin{equation}
v^{ave}_{lk} = \frac{1}{{\left| {\Lambda _{lk}} \right|}}\sum\limits_{\left( {i,j} \right) \in \Lambda _{lk}} {{f_{l-1}}\left( {i,j,c} \right)} ,c \in \left[ {1,C_{l}} \right], k \in \left[ {1,K_l} \right],
\label{eq:gsa_ave}
\end{equation}where $v_{lk} \in {R^{C_{l}}}$ is the category-wise feature of the $k^{th}$ category at Level $l$, $K_l$ is the number of categories at Level $l$, $i.e.$, $K_1 = 2$, $K_2 = 5$. $f_{l}$ is the feature map at Level $l$ and we define $f_0 = f$. $\Lambda _{lk}$ is the index set (category mask) denoting the pixel index corresponding to the $k^{th}$ category at Level $l$ in the final prediction. To enhance the feature representation, we also adopt max pooling instead of average pooling as in Eq.~\eqref{eq:gsa_ave} to calculate the category-wise features, $i.e.$,
\begin{equation}
v_{lk}^{max} = \mathop {\max }\limits_{\left( {i,j} \right) \in {\Lambda _{lk}}} {f_{l - 1}}\left( {i,j,c} \right),c \in \left[ {1,C_l} \right], k \in \left[ {1,K_l} \right].
\label{eq:gsa_max}
\end{equation}
Then, we concatenate $v^{ave}_{lk}$ and $v^{max}_{lk}$ as the aggregated node feature:

${v_{lk}} = concat\left( {v_{lk}^{ave},v_{lk}^{\max }} \right).$
It is noteworthy that obtaining the predictions at different levels is straightforward by referring the definition in the above section. This process is illustrated in the left part in Figure~\ref{fig:gpmLevel2}. Consequently, we can construct a graph representation ${G_l} = \left( {{V_l},{E_l}} \right)$ at each level. $V_l$ and $E_l$ are the node set and edge set in the graph. Accordingly, we use $v_{lk}$ denotes the $k^{th}$ node feature in $G_l$.

\subsubsection{Graph Context Reasoning}
\label{subsec:GCR}
As the nodes in the aforementioned graph corresponds to the specific human body parts, they are correlated with each other. For example, the head, arms, and legs are connected to the torso. Usually, for an upright human body, the legs are on the bottom part, the torso is on the middle part, and the head is on the top part. To model the correlations within nodes, a naive choice is graph convolutional neural network \cite{li2018deeper,gong2019graphonomy}. However, we argue that it is not trivial to define the adjacent matrix. For example, one node may depend on the other even if they are not connected. In this case, though using graph convolution for multiple times can propagate information from one node to its target node, it is inefficient to model such a long-range dependency. To address this issue, we adopt self-attention to model the correlations between context nodes, where the dependency between any two nodes is learnable as the attention weight. Mathematically, it can be formulated as:
\begin{equation}
{a_l} = soft\max \left( \left( {v_l}{Q_{l1}} \right) \cdot {\left( {v_l}{Q_{l2}} \right)}^T \right),
\label{eq:attentionscore}
\end{equation}

where ${v_l} \in {R^{{K_l} \times {C_l}}}$ is the above concatenated node feature, ${Q_{l1}}$ (or ${Q_{l2}}) \in {R^{{C_l} \times \frac{C_l}{8}}}$ denotes the weight matrix in a bottleneck 1*1 convolutional layer, which projects ${v_l}$ into a low-dimensional feature. Then we get the attention vector ${a_l} \in {R^{{K_l} \times {K_l}}}$ by multiplying them and a softmax. Then, the self-attended node feature can be calculated as follows: 
\begin{equation}
v_l^{att} = {a_l}{v_l}.
\label{eq:attentedfeature}
\end{equation}
Finally, the refined node feature is obtained by adding the node feature and self-attended node feature together following the residual learning strategy, $i.e.$,
\begin{equation}
{v^{gcr}_{l}} = {v_l} + v_l^{att},
\label{eq:refinedfeature}
\end{equation}
where $v^{gcr}_l$ is the refined node feature. The above refinement can be carried out several times (3 in this paper) where the refined node feature is used as the input of the latter refinement again. In this way, we can model the correlations within context nodes and call this process as \emph{graph context reasoning} as illustrated in Figure~\ref{fig:gpmLevel2}. For simplicity, we reuse the notation $v^{gcr}_l$ without causing ambiguity.

\subsubsection{Graph Semantics Distribution}
\label{subsec:GSD}
After obtaining the refined node feature, we can re-project it onto the feature map $f_{l-1}$ and fuse them. It can be formulated as:
\begin{equation}
{f_{l}}\left( {i,j,c} \right) = {f_{l-1}}\left( {i,j,c} \right) + \sum\limits_{k = 1}^{{K_l}} {v_l^{gcr}\left( k, c \right)\delta \left( {i,j,k} \right)},
\label{eq:gsd}
\end{equation}
where $\delta \left(  \cdot  \right)$ is an indicator function, $i.e.$, $\delta \left( {i,j,k} \right) = 1$ if $\left( {i,j} \right) \in {\Lambda _{lk}}$, else 0. This process distributes the refined semantical node features to corresponding pixels under the guidance of category masks. So, we call it \emph{graph semantics distribution} as illustrated in the right part in Figure~\ref{fig:gpmLevel2}.

\subsubsection{Progressive Hierarchical Refinement}
\label{subsec:PHR}
As we have constructed a graph representation at each level and defined context feature refinement on it, we can cascade them sequentially to make a graph pyramid. Figure~\ref{fig:grapyml}(a) illustrate the graph pyramid module. As can be seen, we start aggregating, refining and distributing the graph node features at the coarse level. Then, we move to the next level and progressively refine the learned features step by step. Finally, we concatenate the refined feature at each level with the initial one as the input of the final prediction layer. Mathematically, it can be formulated as:
\begin{equation}
\widehat f = concat\left( {{f_l}\left| {l = 0,1,2,3} \right.} \right),
\label{eq:concat}
\end{equation}
\begin{equation}
\widehat y = \widehat p\left( {\widehat f} \right),
\label{eq:predictionGPM}
\end{equation}
where $\widehat f$ denotes the fused feature, $\widehat p\left(  \cdot  \right)$ and $\widehat y$ denotes the final prediction layer and the final prediction at the GPM branch, respectively. They have similar meanings and dimensions as in Eq.~\eqref{eq:rawprediction}.

\subsubsection{Training objective}
\label{subsec:loss}
Once we obtain the predictions from the main branch ($y$) and the GPM branch ($\widehat y$), we can use the cross-entropy loss as training objectives and employ SGD optimizer to train the whole network. Here, we define the multi-task training objective as follows:

\begin{equation}
\begin{array}{l}
 L = {L_{main}} + {L_{GPM}} \\
 \quad  =  - \sum\limits_{i,j,k} {q\left( {i,j,k} \right)\log \left( {y\left( {i,j,k} \right)} \right)}  \\
 \qquad - \lambda \sum\limits_{i,j,k} {q\left( {i,j,k} \right)\log \left( {\widehat y\left( {i,j,k} \right)} \right)}  \\
 \end{array},
\label{eq:loss}
\end{equation}
where $L_{main}$ and $L_{GPM}$ are the losses in the main branch and the GPM branch, respectively. $\lambda$ is the weight for loss. $q$ is the ground truth label at the finest level.

\subsection{GPM-based Mutual Learning for Cross-dataset Human Parsing}
\label{subsec:grapyml}
As we have mentioned that the annotations are heterogeneous and at different granularities. To learn a robust feature presentation and improve the parsing performance, it is beneficial to leverage the heterogeneous multi-granularity annotations across different datasets. To this end, we propose a novel mutual learning method based on the GPM.

Retrospecting the graph pyramid we have defined above, ground truth category labels at the first two coarse granularity levels can be deduced from the annotations in any of the popular human parsing datasets, such as PASCAL-Person Part dataset, CIHP dataset, and ATR dataset. Therefore, one possible solution is to add three GPM branches for each of the datasets and employ multi-task learning to train the whole network. However, we argue that the categories at the first coarse levels have clear definitions, which are the same among all the three datasets. Based on this observation, we propose a mutual learning method by sharing the first two levels of the GPM and adding three specific branches at the finest level for each dataset, as illustrated in Figure~\ref{fig:grapyml}(b). In this way, the network is divided into two parts: Domain Agnostic Module and Domain Specific Module.
In the DAM, the node features at the first two levels are learned by using all the datasets, which will be more robust and discriminative. In addition, since the coarse granularity nodes are shared by the following three branches, a better coarse granularity node feature representation will benefit the feature learning at the finest level through the proposed progressive refinement in DSM. In this sense, the human body information learned from all the datasets are exchanged via the forward prediction and the backward propagation.

Specifically, the training objective $L_{ml}$ for mutual learning on multiple datasets can be formulated as follows:
\begin{equation}
L_{ml} = \sum\limits_{d = 1}^3 {L_{main}^d + L_{GPM}^d},
\label{eq:lossml}
\end{equation}
where $d$ is the index of the training dataset, $L^d_{main}$ and $L^d_{GPM}$ have the same definition as in Eq.~\eqref{eq:loss}.

\section{Experiments}
In this section, we evaluated Grapy-ML on three datasets including PASCAL-Person Part dataset \cite{chen2014detect}, CIHP \cite{gong2018instance} and ATR \cite{liang2015} from three aspects: quantitative comparison, visual inspection, and ablation study.

\subsection{Datasets and Implementation Details}

\subsubsection{Datasets and Evaluation Metric} PASCAL-Person-Part \cite{chen2014detect} dataset contains 3,535 annotated images distributed to 1,717 for training and 1,818 for testing, among which only 7 categories are labeled including 6 human body parts and the background. 7,700 images with 18 categories are provided in ATR dataset \cite{liang2015}, and  \cite{liang2017interpretable} enriched it with 10000 more annotated images. Different from previous single-person datasets, CIHP dataset \cite{gong2018instance} is a popular multi-person one that has 28,280 images for training, 5,000 images for testing and 5,000 images for validation with 20 categories. Following \cite{chen2018encoder,gong2019graphonomy}, we carried out multi-person human parsing without using the provided instance maps. In this section, we use mean accuracy and mean intersection over union (IoU) as the evaluation matrix.

\subsubsection{Implementation Details}
We adopt DeepLab v3+ \cite{chen2018encoder} with Xception \cite{chollet2017xception} as our backbone network and the output stride was set to 16. All experiments were conducted using two NVIDIA Tesla V100 GPUs with batch size 10. The training images were augmented by a random resize from 0.5 to 2, $512 \times 512$ cropping and horizontal flipping. We used a poly learning policy during training \cite{chen2018encoder}. $\lambda$ in Eq. \eqref{eq:loss} was set to 1. Two models named GPM and Grapy-ML were trained, where GPM was trained on a single dataset without using mutual learning. Specifically, we first pretrained the backbone network for 100 epochs with an initial learning rate of 0.007 for the GPM model. Note that the pretrain stage was necessary since we needed a good initial prediction to calculate graph node features. Then the whole model was trained for another 200, 100 and 100 epochs on PASCAL, CIHP, and ATR dataset by decreasing the learning rate to 0.0007. When training the Grapy-ML model, the pretrain stage was conducted on the combination of all three datasets with the same hyper-parameters. Then, we adopt a fine-tuning strategy to adapt the model to the target dataset for another 30 epochs. During testing for both models, we averaged all predictions from horizontal flipped and multi-scale inputs as the final prediction by referring to Graphonomy\cite{gong2019graphonomy}.
\subsection{Main Results}
\begin{table}
\centering
\resizebox{0.47\textwidth}{!}
{\begin{tabular}{c|c}
\noalign{\smallskip} \hline \noalign{\smallskip}
Methods & mIoU \\
\hline
PGN \cite{gong2018instance} & 68.4 \\
Bilinski \cite{bilinski2018dense} & 68.6 \\
DeepLab v3+ \cite{chen2018encoder} & 68.6 \\
\textbf{GPM} & \textbf{69.50} \\
\hline
Graphonomy (Universal) \cite{gong2019graphonomy} & 69.12 \\
Graphonomy (Transfer) \cite{gong2019graphonomy} & 71.14 \\
\textbf{Grapy-ML} & \textbf{71.65} \\
\hline
\end{tabular}}
\caption{Comparison on PASCAL-Person-Part Dataset.}
\label{tab:1}
\end{table}

\begin{table}
\centering
\resizebox{0.47\textwidth}{!}
{\begin{tabular}{c|c|c}
\noalign{\smallskip} \hline \noalign{\smallskip}
Methods & Mean Accuracy & mIoU \\
\hline
JPPNet \cite{liang2018look} & --- & 54.45 \\
DeepLab v3+ \cite{chen2018encoder} & 0.8407 & 76.52 \\
\textbf{GPM} & \textbf{0.8444} & \textbf{76.97} \\
\hline
Graphonomy (Universal) \cite{gong2019graphonomy} & 0.8398 & 76.35 \\
\textbf{Grapy-ML} & \textbf{0.8522} & \textbf{77.88} \\
\hline
\end{tabular}}
\caption{Comparison on ATR Dataset.}
\label{tab:2}
\end{table}

\begin{table}
\centering
\resizebox{0.47\textwidth}{!}
{\begin{tabular}{c|c|c}
\noalign{\smallskip} \hline \noalign{\smallskip}
Methods & Mean Accuracy & mIoU \\
\hline
PGN \cite{gong2018instance} & 64.22 & 55.80 \\
DeepLab v3+ \cite{chen2018encoder} & 67.69 & 58.95 \\
\textbf{GPM} & \textbf{68.95} & \textbf{60.36}\\
\hline
Graphonomy (Universal) \cite{gong2019graphonomy} & 65.73 & 57.78 \\
Graphonomy (Transfer) \cite{gong2019graphonomy} & 66.65 & 58.58 \\
\textbf{Grapy-ML} & \textbf{68.97} & \textbf{60.60}\\
\hline
\end{tabular}
}
\caption{Comparison on CIHP Dataset.}
\label{tab:3}
\end{table}

\begin{figure*}[ht]
\centering
\includegraphics[width=0.99\linewidth]{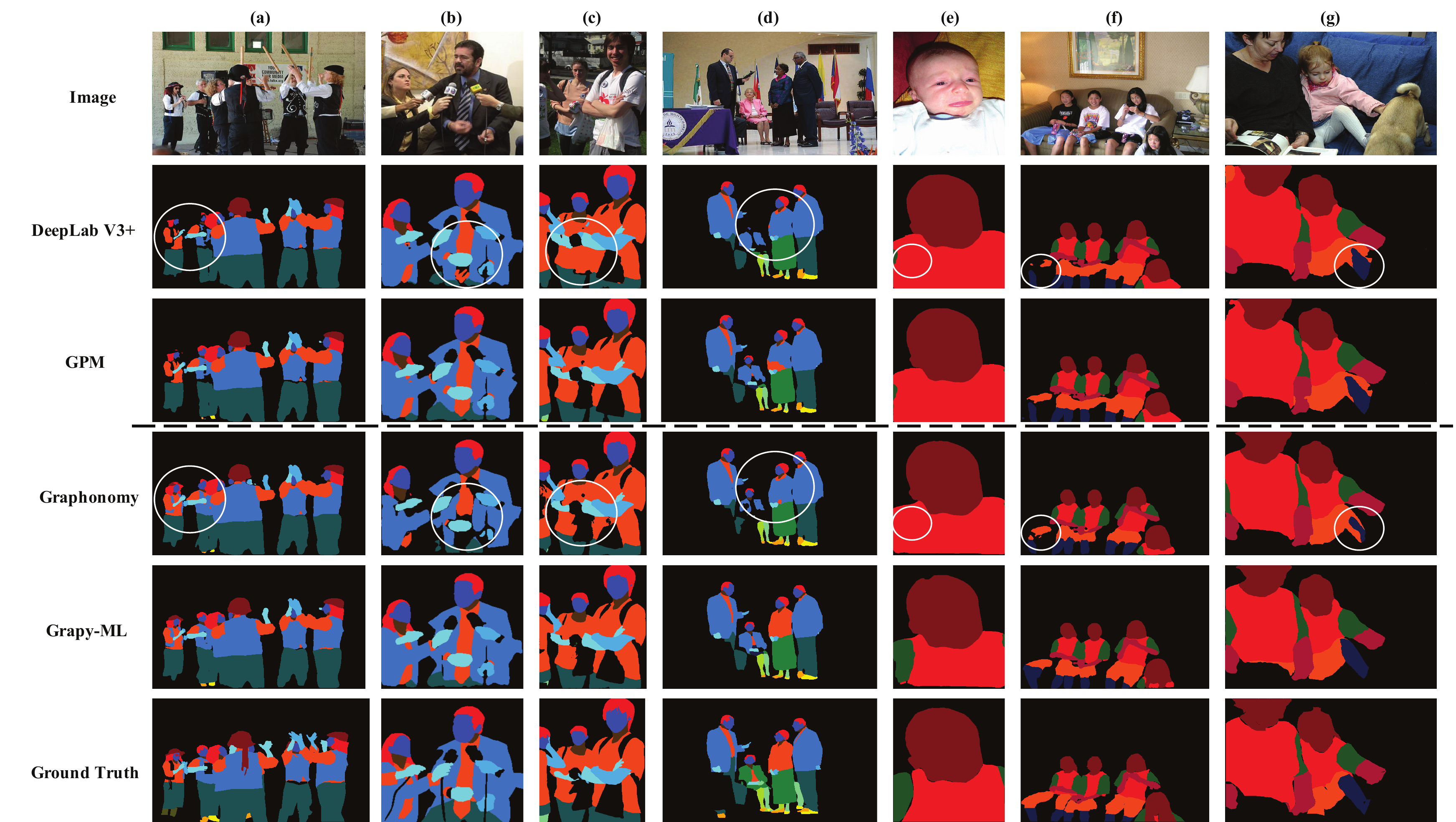}
\caption{Visualized results from the validation set of CIHP (a, b, c, d) and PASCAL-Person-Part (e, f, g).}
\label{fig:visualization}
\end{figure*}

\subsubsection{Quantitative Results}
We evaluate the performance of our model in this section with the best performance highlighted in bold. For fairness, the results of the GPM model are compared with previous methods trained on a single dataset, and Grapy-ML model is with those on multi-datasets.

The results of PASCAL-Person-Part dataset are shown in Table~\ref{tab:1}. From the table, our proposed GPM model achieves the best performance with a mIoU 69.5, proving that our pyramid structure has a good ability to model the relations among body parts. When incorporating multi-datasets, the performance is further improved to 71.65 mIoU. As can be seen, our Grapy-ML model outperforms Graphonomy (Transfer) \cite{gong2019graphonomy} method by a healthy margin, $i.e.$, 0.5 mIoU. The improvement of mutual learning comes from two aspects. Firstly, PASCAL-Person-Part dataset itself is relatively small, within which only 1,717 training data are provided. Thus abundant extra annotated samples in the other two datasets help enhance the generalization ability. Secondly, the coarse-grained category information in the other datasets (ATR dataset and CIHP dataset) boosts the fine-grained parsing quality in all three datasets through the shared two coarse-grained levels, which demonstrates the superiority of our Grapy-ML model.

However, comparing to Graphonomy, such performance gain by Grapy-ML is not as significant as ATR and CIHP. Since Graphonomy models the implicit relationship between cross-dataset features by utilizing inter-dataset transferring matrix.
In contrast to it, the proposed Grapy-ML learns a unified feature representation at the coarse levels from all the datasets under the explicit guidance of the hierarchical category-wise masks. Then, by sharing the coarse-level features, it further refines the fine-grained features at level 3 accordingly. Therefore, Grapy-ML achieves larger gains on ATR and CIHP with fine-grained annotations compared with the Graphonomy model.

We present the mean accuracy and mIoU on ATR dataset in Table~\ref{tab:2}. According to the table, our proposed model obtains the best performance both on the single dataset and on multi-human datasets settings. Specifically, the mean accuracy and mIoU reaches 84.44 and 76.97 respectively on the single dataset and increases to 85.22 and 77.88 when incorporating mutual learning strategy. It is noted that the granularity levels of labels in ATR dataset (18 labels in total) are different from the ones in other datasets. Nevertheless, Grapy-ML achieves a gain of 1 mIoU which again validates the effectiveness of the proposed GPM-based mutual learning. The graph pyramid module successfully learns more robust coarse-level features, which further improve the parsing performance at the fine-granularity level through progressive refinement.

It is worth mentioning that the GPM model only improves the performance by 0.45 compared with the baseline, which is less than the improvement on the other two datasets. It is because parsing a single person with limited pose and size diversity from ATR is much easier than its counterpart from PASCAL and CIHP. Therefore, the baseline Deeplab v3+ have achieved fairly good performance and the improvement by GPM is marginal.

We report the results on CIHP dataset in Table~\ref{tab:3}. The proposed model still outperforms representative state-of-the-art methods significantly, with the mean accuracy reaching 68.97 and mIoU reaching 60.60, proving the superiority of the proposed graph pyramid module. However, the improvement after using mutual learning is marginal. It owes to the large volume of CIHP dataset that extra comparable amount of data contributes marginally (28k in CIHP dataset v.s. 21k in PASCAL-Person Part + ATR datasets), especially when there exists a potential domain gap. We anticipate that further improvement can be achieved by utilizing the instance-level annotations and domain adaptation techniques. We leave it as our future work.

\subsubsection{Visual Inspection}
Some visual results on CIHP and PASCAL-Person-Part dataset from the GPM model, the Grapy-ML model and representative models are visualized in Figure ~\ref{fig:visualization}. As can be seen, our Grapy-ML model produces parsing results with higher precision and consistency. For example, part of the tie enclosed by the white circle in (a) is occluded by the man's right hand and is divided into two discontinuous parts. In this case, the other two methods fail to capture the lower part. However, owing to the graph semantics aggregation and progressive graph context reasoning which effectively explicitly models body-parts correlations, our Grapy-ML successfully eliminates the side-effects of occlusion and correctly segments out the tie. In (b) and (c), the similar colors and textures lead to ambiguity for the other two methods, $i.e.$, the woman's dark blue clothes and the man's black coat, the greyish-white bag and the man's white T-shirt, while this issue is addressed by the coarse-to-fine learning strategy in the proposed model. For (b), the pixels of the upper clothes belong to the torso category at Level 2. Then at the finest level, they are further associated with some fine-grained categories including upper clothes, dress, and coat. Thus the finest-level network in our model can focus on the inner difference of similar items, and the whole parsing task is not as hard as the single-pass methods without using hierarchical multi-granularity refinement. Consequently, the clothes and coats in (b) are segmented appropriately. In (c), the left arm, bag, background, and T-shirt torso are distinguished correctly. The parsing result seems marginal more smooth without the black background holes, especially for the pixels of the upper clothes (orange).
Besides, the benefit of the coarse-to-fine strategy appears in small size items, as shown in (d), where nearly all parts including arms, hair, hats and even feet are segmented into the right categories. The same conclusion can also be drawn from (e), (f), and (g).

We investigate the effectiveness of key components in our Grapy-ML model on CIHP dataset in Table~\ref{tab:ablation}. Firstly, we examine the graph context reasoning methods by comparing graph convolutions and self-attention. As can be seen, employing GCN for reasoning after GSA leads to a significant margin of 0.9 mIoU over the strong Deeplab v3+ baseline. For the proposed GCR using self-attention with both average and max feature, a gain of 1.2 mIoU is achieved, proving the effectiveness of our proposed GCR module. Then the result reaches 60.36 mIoU after employing the complete multi-granularity graph pyramid module. However, when using GCR with only average feature, GPM obtains an mIoU of 60.05 (59.89 for using only max feature). The two results are marginally inferior to the one using both features. Moreover, the proposed GPM-based mutual learning method further improve the performance to 60.6 mIoU as shown in the last row, which set a new state-of-the-art result on this benchmark. It demonstrates that the Grapy-ML takes the advantages of the abundant multi-granularity annotations from multiple datasets via the efficient graph pyramid representation.

\begin{table}[hbt]
\centering
\resizebox{0.47\textwidth}{!}{%
\begin{tabular}{c|c|c|c|c|c}
\hline
\multicolumn{4}{c|}{GPM} & \multirow{3}{*}{Mutual Learning} & \multirow{3}{*}{mIoU} \\ \cline{1-4}
\multicolumn{3}{c|}{Level 3} & \multirow{2}{*}{Level 1 \& Level 2} &  &  \\ \cline{1-3}
GCN & GCR-ave & GCR-max &  &  &  \\ \hline
 &  &  &  &  & 58.95 \\ \hline
\checkmark &  &  &  &  & 59.81 \\
 & \checkmark & \checkmark &  &  & 60.10 \\
 & \checkmark &  & \checkmark &  & 60.05 \\
 &  & \checkmark & \checkmark &  & 59.89 \\
 & \checkmark & \checkmark & \checkmark &  & 60.36 \\
 & \checkmark & \checkmark & \checkmark & \checkmark & 60.60 \\ \hline
\end{tabular}%
}
\caption{Human parsing ablation study in CIHP dataset. Level 1$\sim$3 denote the three levels of GPM. GCN represents using graph convolutions for graph context reasoning in the proposed GPM. GCR represents using self-attention for graph context reasoning in the proposed GPM.}
\label{tab:ablation}
\end{table}

\subsection{Conclusion}
In this work, we introduce a novel graph pyramid mutual learning method named Grapy-ML to address the cross-dataset human parsing problem. Grapy-ML is built on a three-level graph pyramid module by inheriting from a specifically defined multi-granularity lexical pyramid. Within the GPM, features are progressively refined through graph semantics aggregation, graph context reasoning, and graph semantics distribution, where self-attention is explored to model the correlations between graph node features at different granularities. Furthermore, we adopt mutual learning on GPM by sharing the coarse-level graphs across different datasets, which efficiently makes use of the heterogeneous multi-granularity annotations to learn robust features and improve the parsing performance. The experiments demonstrate that Grapy-ML produces satisfying parsing results on several popular human parsing datasets and outperforms state-of-the-art models.

\subsection{Acknowledgement}
This work was supported by the Australian Research Council Projects FL-170100117, DP-180103424, IH-180100002.

\bibliographystyle{aaai}

{
\small
\bibliography{2317-References}

\begin{thebibliography}{}

\bibitem[\protect\citeauthoryear{Bilinski and
  Prisacariu}{2018}]{bilinski2018dense}
Bilinski, P., and Prisacariu, V.
\newblock 2018.
\newblock Dense decoder shortcut connections for single-pass semantic
  segmentation.
\newblock In {\em Proceedings of the IEEE Conference on Computer Vision and
  Pattern Recognition},  6596--6605.

\bibitem[\protect\citeauthoryear{Chen \bgroup et al\mbox.\egroup
  }{2014}]{chen2014detect}
Chen, X.; Mottaghi, R.; Liu, X.; Fidler, S.; Urtasun, R.; and Yuille, A.
\newblock 2014.
\newblock Detect what you can: Detecting and representing objects using
  holistic models and body parts.
\newblock In {\em Proceedings of the IEEE Conference on Computer Vision and
  Pattern Recognition},  1971--1978.

\bibitem[\protect\citeauthoryear{Chen \bgroup et al\mbox.\egroup
  }{2018}]{chen2018encoder}
Chen, L.-C.; Zhu, Y.; Papandreou, G.; Schroff, F.; and Adam, H.
\newblock 2018.
\newblock Encoder-decoder with atrous separable convolution for semantic image
  segmentation.
\newblock In {\em Proceedings of the European Conference on Computer Vision},
  801--818.

\bibitem[\protect\citeauthoryear{Chollet}{2017}]{chollet2017xception}
Chollet, F.
\newblock 2017.
\newblock Xception: Deep learning with depthwise separable convolutions.
\newblock In {\em Proceedings of the IEEE conference on computer vision and
  pattern recognition},  1251--1258.

\bibitem[\protect\citeauthoryear{Dai, He, and Sun}{2016}]{dai2016instance}
Dai, J.; He, K.; and Sun, J.
\newblock 2016.
\newblock Instance-aware semantic segmentation via multi-task network cascades.
\newblock In {\em Proceedings of the IEEE Conference on Computer Vision and
  Pattern Recognition},  3150--3158.

\bibitem[\protect\citeauthoryear{Dong \bgroup et al\mbox.\egroup
  }{2014}]{dong2014towards}
Dong, J.; Chen, Q.; Shen, X.; Yang, J.; and Yan, S.
\newblock 2014.
\newblock Towards unified human parsing and pose estimation.
\newblock In {\em Proceedings of the IEEE Conference on Computer Vision and
  Pattern Recognition},  843--850.

\bibitem[\protect\citeauthoryear{Gong \bgroup et al\mbox.\egroup
  }{2017}]{gong2017look}
Gong, K.; Liang, X.; Zhang, D.; Shen, X.; and Lin, L.
\newblock 2017.
\newblock Look into person: Self-supervised structure-sensitive learning and a
  new benchmark for human parsing.
\newblock In {\em Proceedings of the IEEE Conference on Computer Vision and
  Pattern Recognition},  932--940.

\bibitem[\protect\citeauthoryear{Gong \bgroup et al\mbox.\egroup
  }{2018}]{gong2018instance}
Gong, K.; Liang, X.; Li, Y.; Chen, Y.; Yang, M.; and Lin, L.
\newblock 2018.
\newblock Instance-level human parsing via part grouping network.
\newblock In {\em Proceedings of the European Conference on Computer Vision},
  770--785.

\bibitem[\protect\citeauthoryear{Gong \bgroup et al\mbox.\egroup
  }{2019}]{gong2019graphonomy}
Gong, K.; Gao, Y.; Liang, X.; Shen, X.; Wang, M.; and Lin, L.
\newblock 2019.
\newblock Graphonomy: Universal human parsing via graph transfer learning.
\newblock In {\em Proceedings of the IEEE Conference on Computer Vision and
  Pattern Recognition},  7450--7459.

\bibitem[\protect\citeauthoryear{He \bgroup et al\mbox.\egroup
  }{2017}]{he2017mask}
He, K.; Gkioxari, G.; Doll{\'a}r, P.; and Girshick, R.
\newblock 2017.
\newblock Mask r-cnn.
\newblock In {\em Proceedings of the IEEE International Conference on Computer
  Vision},  2961--2969.

\bibitem[\protect\citeauthoryear{Huang, Gong, and Tao}{2017}]{huang2017coarse}
Huang, S.; Gong, M.; and Tao, D.
\newblock 2017.
\newblock A coarse-fine network for keypoint localization.
\newblock In {\em Proceedings of the IEEE International Conference on Computer
  Vision},  3028--3037.

\bibitem[\protect\citeauthoryear{Li \bgroup et al\mbox.\egroup
  }{2017}]{li2017multiple}
Li, J.; Zhao, J.; Wei, Y.; Lang, C.; Li, Y.; Sim, T.; Yan, S.; and Feng, J.
\newblock 2017.
\newblock Multiple-human parsing in the wild.
\newblock {\em arXiv preprint arXiv:1705.07206}.

\bibitem[\protect\citeauthoryear{Li, Han, and Wu}{2018}]{li2018deeper}
Li, Q.; Han, Z.; and Wu, X.-M.
\newblock 2018.
\newblock Deeper insights into graph convolutional networks for semi-supervised
  learning.
\newblock In {\em Proceedings of the AAAI Conference on Artificial
  Intelligence}.

\bibitem[\protect\citeauthoryear{Liang \bgroup et al\mbox.\egroup
  }{2015a}]{liang2015}
Liang, X.; Liu, S.; Shen, X.; Yang, J.; Liu, L.; Dong, J.; Lin, L.; and Yan, S.
\newblock 2015a.
\newblock Deep human parsing with active template regression.
\newblock {\em IEEE transactions on pattern analysis and machine intelligence}
  37(12):2402--2414.

\bibitem[\protect\citeauthoryear{Liang \bgroup et al\mbox.\egroup
  }{2015b}]{liang2015human}
Liang, X.; Xu, C.; Shen, X.; Yang, J.; Liu, S.; Tang, J.; Lin, L.; and Yan, S.
\newblock 2015b.
\newblock Human parsing with contextualized convolutional neural network.
\newblock In {\em Proceedings of the IEEE International Conference on Computer
  Vision},  1386--1394.

\bibitem[\protect\citeauthoryear{Liang \bgroup et al\mbox.\egroup
  }{2016a}]{liang2016semantic2}
Liang, X.; Shen, X.; Feng, J.; Lin, L.; and Yan, S.
\newblock 2016a.
\newblock Semantic object parsing with graph lstm.
\newblock In {\em Proceedings of the European Conference on Computer Vision},
  125--143.
\newblock Springer.

\bibitem[\protect\citeauthoryear{Liang \bgroup et al\mbox.\egroup
  }{2016b}]{liang2016semantic}
Liang, X.; Shen, X.; Xiang, D.; Feng, J.; Lin, L.; and Yan, S.
\newblock 2016b.
\newblock Semantic object parsing with local-global long short-term memory.
\newblock In {\em Proceedings of the IEEE Conference on Computer Vision and
  Pattern Recognition},  3185--3193.

\bibitem[\protect\citeauthoryear{Liang \bgroup et al\mbox.\egroup
  }{2017}]{liang2017interpretable}
Liang, X.; Lin, L.; Shen, X.; Feng, J.; Yan, S.; and Xing, E.~P.
\newblock 2017.
\newblock Interpretable structure-evolving lstm.
\newblock In {\em Proceedings of the IEEE Conference on Computer Vision and
  Pattern Recognition},  1010--1019.

\bibitem[\protect\citeauthoryear{Liang \bgroup et al\mbox.\egroup
  }{2018}]{liang2018look}
Liang, X.; Gong, K.; Shen, X.; and Lin, L.
\newblock 2018.
\newblock Look into person: Joint body parsing \& pose estimation network and a
  new benchmark.
\newblock {\em IEEE transactions on pattern analysis and machine intelligence}
  41(4):871--885.

\bibitem[\protect\citeauthoryear{Luo \bgroup et al\mbox.\egroup
  }{2018}]{luo2018macro}
Luo, Y.; Zheng, Z.; Zheng, L.; Guan, T.; Yu, J.; and Yang, Y.
\newblock 2018.
\newblock Macro-micro adversarial network for human parsing.
\newblock In {\em Proceedings of the European Conference on Computer Vision},
  418--434.

\bibitem[\protect\citeauthoryear{Nie, Feng, and Yan}{2018}]{nie2018mutual}
Nie, X.; Feng, J.; and Yan, S.
\newblock 2018.
\newblock Mutual learning to adapt for joint human parsing and pose estimation.
\newblock In {\em Proceedings of the European Conference on Computer Vision},
  502--517.

\bibitem[\protect\citeauthoryear{Ruan \bgroup et al\mbox.\egroup
  }{2019}]{ruan2019devil}
Ruan, T.; Liu, T.; Huang, Z.; Wei, Y.; Wei, S.; and Zhao, Y.
\newblock 2019.
\newblock Devil in the details: Towards accurate single and multiple human
  parsing.
\newblock In {\em Proceedings of the AAAI Conference on Artificial
  Intelligence}, volume~33,  4814--4821.

\bibitem[\protect\citeauthoryear{Xiao \bgroup et al\mbox.\egroup
  }{2018}]{xiao2018unified}
Xiao, T.; Liu, Y.; Zhou, B.; Jiang, Y.; and Sun, J.
\newblock 2018.
\newblock Unified perceptual parsing for scene understanding.
\newblock In {\em Proceedings of the European Conference on Computer Vision},
  418--434.

\bibitem[\protect\citeauthoryear{Yang \bgroup et al\mbox.\egroup
  }{2019}]{yang2019parsing}
Yang, L.; Song, Q.; Wang, Z.; and Jiang, M.
\newblock 2019.
\newblock Parsing r-cnn for instance-level human analysis.
\newblock In {\em Proceedings of the IEEE Conference on Computer Vision and
  Pattern Recognition},  364--373.

\bibitem[\protect\citeauthoryear{Zhao \bgroup et al\mbox.\egroup
  }{2018}]{zhao2018understanding}
Zhao, J.; Li, J.; Cheng, Y.; Sim, T.; Yan, S.; and Feng, J.
\newblock 2018.
\newblock Understanding humans in crowded scenes: Deep nested adversarial
  learning and a new benchmark for multi-human parsing.
\newblock In {\em 2018 ACM Multimedia Conference on Multimedia Conference},
  792--800.
\newblock ACM.

\bibitem[\protect\citeauthoryear{Zhu \bgroup et al\mbox.\egroup
  }{2018}]{zhu2018progressive}
Zhu, B.; Chen, Y.; Tang, M.; and Wang, J.
\newblock 2018.
\newblock Progressive cognitive human parsing.
\newblock In {\em Proceedings of the AAAI Conference on Artificial
  Intelligence},  7607--7614.

\end{thebibliography}
}

\end{document}